\documentclass[10pt,twocolumn,letterpaper]{article}

\usepackage[pagenumbers]{iccv} % To force page numbers, e.g. for an arXiv version

\definecolor{iccvblue}{rgb}{0.21,0.49,0.74}
\usepackage[pagebackref,breaklinks,colorlinks,allcolors=iccvblue]{hyperref}
\usepackage{url}
\usepackage{booktabs}
\usepackage{graphicx}
\usepackage{multirow}
\usepackage{subcaption}
\usepackage{wrapfig}
\usepackage{adjustbox}
\usepackage{enumitem}
\usepackage{algorithm}
\usepackage{algorithmic}
\usepackage{setspace}
\usepackage{bm}
\usepackage{makecell}
\usepackage{caption}
\usepackage{subcaption}
\usepackage{comment}

\usepackage{multicol}
\usepackage{threeparttable}
\usepackage{amsfonts}       % blackboard math symbols
\usepackage{amssymb}     
\usepackage{amsmath}     
\usepackage{nicefrac}       % compact symbols for 1/2, etc.
\usepackage{microtype}      % microtypography
\usepackage{bm}     
\usepackage{xfrac} 
\usepackage{array}
\usepackage{colortbl}

\usepackage{epsfig}
\usepackage{float}
\newcommand{\PreserveBackslash}[1]{\let\temp=\\#1\let\\=\temp}
\newcolumntype{C}[1]{>{\PreserveBackslash\centering}p{#1}}
\newcolumntype{R}[1]{>{\PreserveBackslash\raggedleft}p{#1}}
\newcolumntype{L}[1]{>{\PreserveBackslash\raggedright}p{#1}}

\definecolor{Gray}{rgb}{0.95, 0.95, 0.95}
\def\paperID{717} % *** Enter the Paper ID here
\def\confName{ICCV}
\def\confYear{2025}

\title{Small Clips, Big Gains: Learning Long-Range Refocused Temporal Information for Video Super-Resolution}

\author{\hspace{-0.5cm}
Xingyu Zhou\quad Wei Long\quad
Jingbo Lu\quad  Shiyin Jiang\quad  Weiyi You\quad  Haifeng Wu\quad  Shuhang Gu\thanks{Corresponding author.}\\%\Letter
\hspace{-0.5cm}
University of Electronic Science and Technology of China \hspace{0pt}\\
\hspace{-0.5cm}
{\tt\small \{xy.chous526, shuhanggu\}@gmail.com}\\
\small \url{https://github.com/LabShuHangGU/LRTI-VSR}}

\begin{document}
\maketitle
\begin{abstract}
Video super-resolution (VSR) can achieve better performance compared to single image super-resolution by additionally leveraging temporal information. 
In particular, the recurrent-based VSR model exploits long-range temporal information during inference and achieves superior detail restoration.
However, effectively learning these long-term dependencies within long videos remains a key challenge.
To address this, we propose LRTI-VSR, a novel training framework for recurrent VSR that efficiently leverages \textbf{L}ong-Range \textbf{R}efocused \textbf{T}emporal \textbf{I}nformation. 
Our framework includes a generic training strategy that utilizes temporal propagation features from long video clips while training on shorter video clips.
Additionally, we introduce a refocused intra\&inter-frame transformer block which allows the VSR model to selectively prioritize useful temporal information through its attention module while further improving inter-frame information utilization in the FFN module.
We evaluate LRTI-VSR on both CNN and transformer-based VSR architectures, conducting extensive ablation studies to validate the contribution of each component. 
Experiments on long-video test sets demonstrate that LRTI-VSR achieves state-of-the-art performance while maintaining training and computational efficiency.
\end{abstract}    
\section{Introduction}
\label{sec:intro}
Unlike single image super-resolution (SISR) which relies solely on intra-frame information to estimate missing details, video super-resolution (VSR) additionally leverages temporal information to reconstruct high-resolution frames. 
This key distinction allows VSR to achieve a more accurate recovery of the current frame by exploiting temporal redundancy within video sequences.
Driven by this benefit, the effective use of long-range temporal information has emerged as a critical research focus, playing a pivotal role in the success of modern VSR models.

\begin{figure}[!t]
	\begin{center}
    \hspace{-0.46cm}
	\includegraphics[width=0.99\linewidth]{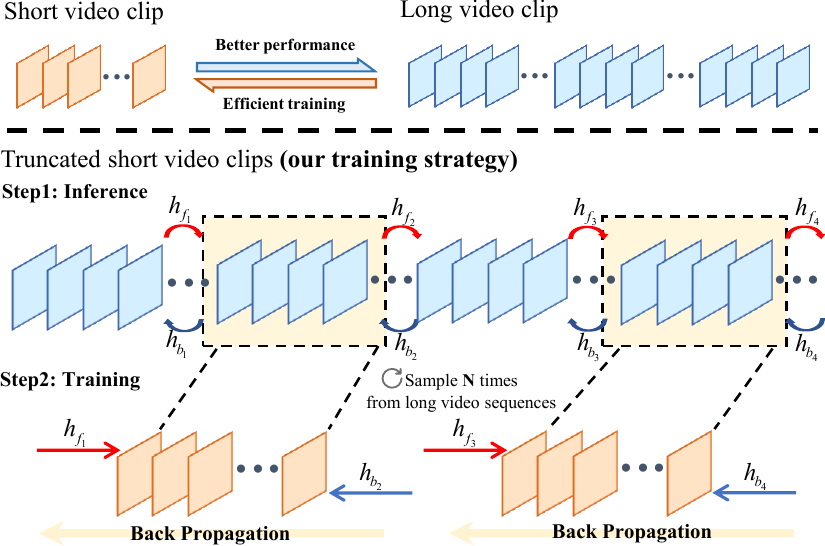}
	\caption{\textbf{The pipline of our proposed training strategy.} Our proposed training strategy can effectively utilize accurate long-range temporal dependencies in long video sequences to assist training while using short video clips for high training efficiency. In this figure,  $\color{red}{\longrightarrow}$$ h_{f}$ means forward propagation hidden state and  $\color{blue}$$\color{blue}{\longleftarrow}$$ h_{b}$ means backward paopagation hidden state in the bidirectional recurrent-based VSR model.}
	\vspace{-0.6cm}
	\label{fig:refill_qual}
	\end{center}
\end{figure}
Early-stage research~\cite{li2020mucan, wang2019edvr, tian2020tdan,liang2022vrt, cao2021video} primarily employed a sliding temporal window strategy, utilizing sophisticated alignment modules and advanced network architectures to generate high-resolution (HR) outputs from multiple low-resolution (LR) inputs.
As the sliding temporal window limits the utilization of temporal information to a fixed window size, recurrent-based VSR models which propagate hidden states across the whole video begin to occupy a major position in the literature of VSR.
In this context, numerous advanced information propagation strategies~\cite{fuoli2019efficient,isobe2020video,chan2021basicvsr,chan2022basicvsr++,isobe2022look,shi2022rethinking,Zhou_2024_CVPR} have been proposed to make more accurate use of long-range temporal information to achieve better VSR performance.
Combined with cutting-edge network architectures, the recurrent-based VSR modeles have significantly advanced the state-of-the-art in VSR field~\cite{chan2022basicvsr++,liang2022vrt,qiu2022learning,liu2022learning,liang2022recurrent, shi2022rethinking,xu2023implicit,Zhou_2024_CVPR}. 

While recurrent-based VSR models have demonstrated strong capabilities in exploiting long-range temporal information, their training becomes increasingly challenging when longer video clips are used as input due to their sequential dependency constraints. 
Recent studies~\cite{chan2021basicvsr,chan2022investigating,liang2022vrt,shi2022rethinking} have found that training recurrent-based VSR models with longer video clips often leads to more accurate VSR results, as longer sequences facilitate the learning of long-range temporal dependencies.
However, as the depth and temporal modeling complexity of the VSR network increases, simply extending the length of training clips becomes highly time-consuming and memory-intensive.
Consequently, state-of-the-art recurrent-based VSR models are typically trained on shorter video clips to balance performance and training efficiency.
This compromise, however, results in models that fail to fully learn accurate long-range temporal dependencies in long video datasets, ultimately degrading performance.
Efficiently capturing long-term information propagation patterns in long video sequences during VSR model training remains a critical open challenge in the VSR field.

In this paper, we provide a feasible solution to this issue with a novel VSR training framework utilizing \textbf{L}ong-Range \textbf{R}efocused \textbf{T}emporal \textbf{I}nformation (LRTI-VSR).
As the pivotal innovation, we introduce a generic training strategy for recurrent-based VSR models that efficiently learns accurate long-term temporal dependencies from long video sequences while maintaining manageable training overhead. 
To be more specific, since training solely on short video clips fails to learn the temporal propagation patterns inherent in long video sequences, we conduct forward and backward propagation in network training with different lengths of video clips.
The forward propagation process is conducted on long video sequences to obtain accurate intermediate hidden states for all the frames, while the backward propagation step is performed on short video clips to facilitate efficient training.
Extensive experiments on our proposed model and a wide range of existing recurrent-based VSR models demonstrate the effectiveness of this strategy, consistently improving performance without modifying the underlying network architectures.
In addition, we rethink the role of aligned redundant temporal propagation hidden states in recovering the current frame and propose a refocused intra- and inter-frame attention structure.
By replacing $\texttt{SoftMax}$ normalization with a sparse refocus activation function $\texttt{ReLU}^2$ in the attention module, our model only selectively prioritizes useful information from previously processed features.
To further enhance inter-frame information utilization, we also integrate the aligned hidden state from the previous frame into the FFN structure through a refocused gate unit.
These improvements enable the proposed refocused intra- and inter-frame transformer block to enhance the LRTI-VSR model's ability to leverage temporal propagation features, establishing a stronger baseline for VSR tasks.
Above all, our contributions can be summarized as follows.

\begin{itemize}%[leftmargin=*]
\item We propose a novel VSR training framework that efficiently leverages accurate long-range temporal information by learning the temporal propagation patterns of long videos and precisely utilizing inter-frame dependencies.
\item We introduce a generic training strategy for recurrent-based VSR models, which enables the effective learning of accurate long-term propagation patterns while training on shorter video clips.
\item We design a refocused intra\&inter-frame transformer block (RITB), which selectively prioritizes useful information from redundant temporal propagation hidden states to enhance current frame recovery.
\item We demonstrate the superiority of our LRTI-VSR framework through extensive comparisons with state-of-the-art VSR models, achieving superior performance while maintaining manageable training overhead. 
\end{itemize}
\vspace{-0.2cm}

\section{Related Work}
\label{gen_inst}
\subsection{Video Super-Resolution}
According to how the temporal information is utilized, all previous deep learning-based VSR methods can be divided into two categories: temporal sliding-window based methods and recurrent based methods.

Temporal sliding-window methods tend to restore a single frame using several neighboring reference frames within a temporal window, such as estimating HR images from multiple input images in a temporal sliding window manner~\cite{li2020mucan, wang2019edvr, cao2021video, liang2022vrt}.
The alignment module plays an essential role in temporal sliding-window based method in modeling the inter-frame relationship.
In the earlier stage, several sliding-window methods~\cite{caballero2017real, liu2017robust, tao2017detail} explicitly estimated the optical flow to align adjacent frames.
Dynamic filters~\cite{jo2018deep}, deformable convolutions~\cite{dai2017deformable, tian2020tdan, wang2019edvr} and attention modules~\cite{Isobe_2020_CVPR, li2020mucan} have been developed to conduct motion compensation implicitly in the feature space.
Although the alignment module allows sliding-window-based VSR networks to better exploit temporal information from adjacent frames, the accessible temporal information is constrained by the window size, limiting the models' ability to utilize data from only a small number of input frames.

Compared with Temporal sliding-window methods that only use short-range temporal information, another category of approaches applies recurrent neural networks to exploit long-range temporal information from more frames.
FRVSR~\cite{sajjadi2018frame} first proposed a recurrent framework that utilizes optical flow to align the previous HR estimation and the current LR input for VSR.
RLSP~\cite{fuoli2019efficient} propagates high-dimensional hidden states instead of the previous HR estimation to better exploit long-term information.
RSDN~\cite{isobe2020video} further extended RLSP~\cite{fuoli2019efficient} by decomposing the LR frames into structure and detail layers and introduced an adaptation module to selectively use the information from hidden states.
BasicVSR~\cite{chan2020basicvsr} utilized bi-directional hidden states, and BasicVSR++~\cite{chan2022basicvsr++} further improved BasicVSR with second-order grid propagation and flow-guided deformable alignment.
Recently, more advanced inter-frame alignment modules (e.g. RVRT~\cite{liang2022recurrent}, PSRT~\cite{shi2022rethinking} and IART~\cite{xu2023implicit}) and computationally efficient network layers (e.g. MIA-VSR~\cite{Zhou_2024_CVPR}) have been proposed to improve the utilization efficiency of inter-frame information.  
Combined with the advanced Transformer architecture, the performance of VSR is improved to a new level.
Our proposed LRTI-VSR model follows the general framework of existing transformer-based recurrent VSR models but leverages long-term dependencies within long video sequences with affordable training overhead.

\subsection{Efficient Modeling of Long Sequences}
How to incorporate long sequence dependencies into training efficiently has always been a key issue in sequence model research.
Training RNNs often relies on the resource-intensive Back-Propagation Through Time~\cite{werbos1990backpropagation} (BPTT) method.
To address its computational challenge, Truncated Back-Propagation Through Time~\cite{williams2013gradient} (TBPTT) was originally designed for recurrent neural networks to model long natural language sequences.
In later stages, researchers proposed unbiased approximations like NoBackTrack~\cite{ollivier2015training} and UORO~\cite{tallec2017unbiased}, which update model parameters online and avoid the memory and computational overhead caused by backpropagation over time, facilitating efficient training of sequence models.
% to reduce memory and computational overhead, facilitating efficient sequence model training. 
ARTBP~\cite{tallec2017unbiasing} utilizes a flexible memory method and compensatory factors to mitigate noise while maintaining accuracy and efficiency for long sequences.
Most recently in the research of large language model, in order to make the Transformer-based sequence model able to use the information of long sequences for training, sparse Transformer~\cite{child2019generating,beltagy2020longformer,ding2023longnet}, compressed memory~\cite{liu2018generating}, KV cache~\cite{shazeer2019fast,dao2022flashattention,kwon2023efficient}, linear Transformer~\cite{katharopoulos2020transformers} and sequence parallelism~\cite{li2021sequence, gu2021efficiently} strategies are usually used for efficient long sequence modeling.
PGT~\cite{pang2021pgt} is the first attempt to introduce the TBPTT strategy into video modeling in high-level vision tasks. 
However, efficiently incorporating long-term information into video restoration model training remains a critical challenge in the field of low-level vision.
To the best of our knowledge, our study is the first work in this field that utilizes long-term propagation patterns of long video sequences to assist training with short video clips, requiring a minor increase in training overhead.

\section{Methodology}
\label{sec:Method}

\subsection{Preliminary}
Given a low-resolution video sequence $\bm{I}^{LR} \in  \mathbb{R}^{T \times H \times W \times 3}$ with $T$ frames, the goal of the VSR model is to reconstruct the corresponding high-resolution video sequence
$\bm{I}^{HR}\in \mathbb{R}^{T \times sH \times sW \times 3}$, where $s$ is the scaling factor and $H$, $W$, $3$ are the height, width and number of channels of the input frames, respectively.

\begin{figure}[!t]
	\begin{center}
		\includegraphics[width=0.46\textwidth]{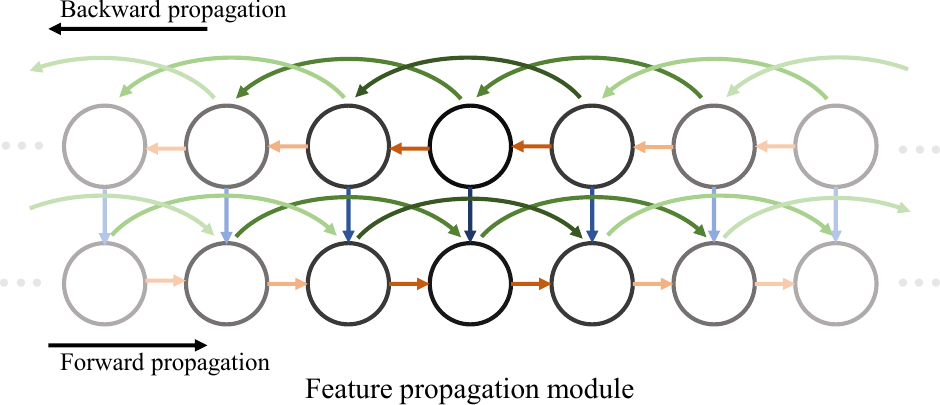}
		\caption{\textbf{Structure of feature propagation module.} The commonly used feature propagation module contains bidirectional feature propagation structure (red solid lines), and second-order connection structure (green solid lines) that leverages the computed hidden states of the previous two frames to recover the current frame. These two components are widely used in existing state-of-the-art VSR models~\cite{chan2022basicvsr++,liang2022recurrent,shi2022rethinking,xu2023implicit,Zhou_2024_CVPR}.}%改
		\vspace{-0.6cm}
		\label{fig:fpm}
	\end{center}
\end{figure}

Current VSR methods can be categorized into sliding-window based and recurrent-based approaches.
Unlike sliding-window based VSR methods which rely on a limited number of adjacent frames, recurrent-based VSR methods exploit long-range temporal information, enabling more efficient and robust restoration during inference.
The structure of temporal information propagation in recurrent-based VSR models (i.e. the feature propagation module) is illustrated in Fig. \ref{fig:fpm}.
As highlighted in previous works on recurrent-based VSR~\cite{fuoli2019efficient,chan2021basicvsr}, the feature propagation module leverages high-dimensional hidden states computed from previous frames to assist in the computation of the current frame feature.
By incorporating longer video sequences during training, the recurrent-based VSR model can more accurately learn long-range temporal propagation patterns, which can significantly improve its performance.

Furthermore, effective modeling of inter-frame information is also crucial for better VSR performance.
To this end, recent studies~\cite{chan2022basicvsr++,shi2022rethinking} have introduced second-order connection structures to extract supplementary temporal information from previous frames, enhancing inter-frame redundancy (as shown in Fig. \ref{fig:fpm}).
Combined with advanced inter-frame alignment techniques~\cite{liang2022recurrent,shi2022rethinking,xu2023implicit} and cross-attention modules~\cite{Zhou_2024_CVPR}, these approaches enable more accurate exploitation of temporal information, further advancing the state-of-the-art in video super-resolution.

\subsection{Efficiently Learning Long-term Dependencies via Short Video Clip Training}
\label{sec:LTI}
As discussed in previous subsection, temporal propagation hidden states play an important role in the performance of recurrent-based video super-resolution models.
Recent studies~\cite{chan2021basicvsr,chan2022investigating,liang2022vrt,shi2022rethinking} have shown that incorporating longer video clips into training leads to better SR results.
However, since recurrent-based VSR methods typically rely on the Back-Propagation Through Time (BPTT)~\cite{werbos1990backpropagation} strategy, simply increasing the length of training video clips makes the training process time-consuming and memory-intensive.
To address this issue, we propose a truncated backpropagation training strategy inspired by TBPTT~\cite{williams2013gradient}. Our approach enables the model to learn temporal propagation patterns from long video sequences while efficiently training on shorter clips.

\textbf{Forward propagation on long video sequences.}
Before training with short video clips as input, the model performs a forward propagation on the corresponding long video sequence $\bm{I}^{LR}=\{x_1, x_2,\dots,x_T\}$ of $T$ frames to obtain the temporal propagation hidden states $H$ for all frames:
\begin{equation}\label{eq:hidden}
    \bm{H} = \texttt{VSR-model}.\texttt{forward}(\bm{I}^{LR}).
\end{equation}
This process follows the original bidirectional recurrent-based VSR inference method, where the initial frame has no additional temporal propagation features as input. 

\textbf{Backpropagation on short video clips.} A short video clip of length $L$ is selected from the long video sequence before training, which can be expressed as:
\begin{equation}\label{eq:clip}
    \bm{I}_{clip}^{LR} = \{\bm{x_t}\}_{t=t_{start}}^{t_{start}+L-1}, t_{start}\sim \texttt{Uniform}(1,T-L+1),
\end{equation}
where $t_{start}$ denotes the starting frame position of the selected short video clip within the long video sequence.
The bidirectional recurrent VSR model then takes the selected video clip along with the corresponding temporal propagation hidden states $\bm{H}_{clip}=\{\bm{h}^{t_{start}-2},\bm{h}^{t_{start}-1},\bm{h}^{t_{start}+L},\bm{h}^{t_{start}+L+1}\}$ before and after starting and ending frames of the clip as input: 
\begin{equation}\label{eq:cliphr}
    \hat{\bm{I}}_{clip}^{HR} = \texttt{VSR-model}(\bm{I}_{clip}^{LR},\bm{H}_{clip}),
\end{equation}
where $\hat{\bm{I}}_{clip}^{HR} $ is the output of the forward propagation process during the VSR model's training. This configuration aligns with the foundational setup of bidirectional second-order grid propagation, a widely adopted paradigm in state-of-the-art VSR frameworks~\cite{chan2022basicvsr++,liang2022recurrent,shi2022rethinking,xu2023implicit,Zhou_2024_CVPR}. 
With leveraging the advanced optical-flow guided feature alignment model~\cite{shi2022rethinking}, our VSR model utilizes the aligned hidden states of previous frames propagated inside and outside the video clip to further assist the restoration of the whole video sequence.

\textbf{Implementation details.} By sampling a fixed number of short clips from each long video for backpropagation, the video super-resolution model gradually converges by leveraging long-term propagation information. The entire workflow of the proposed training strategy is shown in Alg. \ref{alg: training LRTI}. 
\begin{algorithm}
    \caption{Workflow of the proposed training strategy.}
    \label{alg: training LRTI}
    \begin{algorithmic}[1]
    \setstretch{1.10}
        \REQUIRE training VSR-model $\bm{f}_{\bm{\theta}}(\cdot)$
        \REQUIRE Paired training dataset $(\bm{I}^{LR}, \bm{I}^{HR})$, Sample times $N$, Truncated video clip length $L$, Whole video length $T$
        \WHILE {not converged}
            \STATE Sample $\bm{I}^{LR}$
            \STATE  $H =\texttt{VSR-model}.\texttt{forward}(\bm{I}^{LR})$
        % \STATE $\textbf{Stage 1:}$
            \FOR{$n=0,1,\cdots,N-1$}
                \STATE Sample $t_{start}\sim \texttt{Uniform}(1, T-L+1)$
                \STATE $\bm{I}_{clip}^{LR} = \{\bm{x}_t\}_{t=t_{start}}^{t_{start}+L-1}$
                \STATE  $\begin{aligned}
                        \bm{H}_{clip} =\{\bm{h}^{t_{start}-2}, \bm{h}^{t_{start}-1},\\
                        \bm{h}^{t_{start}+L}, \bm{h}^{t_{start}+L+1}\}
                        \end{aligned}$
                \STATE  $\hat{\bm{I}}_{clip}^{HR} = \texttt{VSR-model}(\bm{I}_{clip}^{LR}, \bm{H}_{clip})$
                \STATE $\mathcal{L}_{sr}=\texttt{Charbonnier\_Loss}(\hat{\bm{I}}_{clip}^{HR}, \bm{I}_{clip}^{HR})$
                \STATE $\texttt{VSR-model.backward()}$
        \ENDFOR
        \ENDWHILE
        \RETURN Converged VSR-model $f_\theta(\cdot)$.
    \end{algorithmic}
\end{algorithm}
In our experiments, we intuitively ensure that the product of sampling times and truncation video clips length to be close to the total video sequence length to better balance training efficiency and VSR performance. 
We validate the effectiveness of our training strategy on previous CNN and Transformer-based state-of-the-art VSR models, respectively, and further evaluate the time and memory efficiency of the proposed training strategy.
The detailed experimental setup and results can be seen in Section \ref{sec:experiments}.

\subsection{Refocused Inter- and Intra-Frame Transformer}
\label{sec:ritb}
While utilizing aligned redundant temporal propagation hidden states enhances inter-frame correlation in VSR models, the misalignment caused by the inaccuracy of optical flow induces irrelevant features and unnecessary calculations, negatively impacting performance.
To address this, we propose a \textbf{R}efocused \textbf{I}nter\&intra-Frame \textbf{T}ransformer \textbf{B}lock (\textbf{RITB}), which extends the approach in~\cite{Zhou_2024_CVPR} by directly applying intra- and inter-frame attention to both previously hidden states and current frame features.
Our RITB selectively prioritizes features with positive contributions from inter-frame redundant information while mitigating the influence of irrelevant features arising from the inaccurate alignment module.
Specifically, we first generate Query Tokens from the current frame feature and Key/Value Tokens from both the current frame feature and hidden states of previous frames:
\begin{equation}
\label{eq:QKV}
    \begin{aligned}
    \left\{
    \begin{array}{l}
    \bm{Q}_{m,n}^{t} = \bm{X}_{m,n}^{t}\bm{W}^Q_{m,n},\\
    \bm{K}_{m,n}^{t} =  \left[\bm{h}_{m+1}^{t-1}; \bm{h}_{m+1}^{t-2}; \bm{X}_{m,n}^{t}\right]\bm{W}^K_{m,n},\\
    \bm{V}_{m,n}^{t} =  \left[\bm{h}_{m+1}^{t-1}; \bm{h}_{m+1}^{t-2}; \bm{X}_{m,n}^{t}\right]\bm{W}^V_{m,n},
    \end{array}
    \right.
     \end{aligned}
\end{equation}
Where $\bm{Q}_{m,n}^{t}$, $\bm{K}_{m,n}^{t}$ and $\bm{V}_{m,n}^{t}$ are the Query, Key and Value Tokens generated from the current input feature $\bm{X}_{m,n}^{t}$ and the temporal propagation hidden states $\{\bm{h}_{m+1}^{t-1}, \bm{h}_{m+1}^{t-2}\}$ of the previous frames which are aligned by the feature alignment model; $\bm{W}^Q_{m,n}$, $\bm{W}^K_{m,n}$ and $\bm{W}^V_{m,n}$ are the respective projection matrices; $\bm{m}$ and $\bm{n}$ denote the $\bm{m}$-th feature propagation module and the $\bm{n}$-th RITB block within the feature propagation module, respectively.

\begin{figure}[!t]
	\begin{center}
		\includegraphics[width=0.46\textwidth]{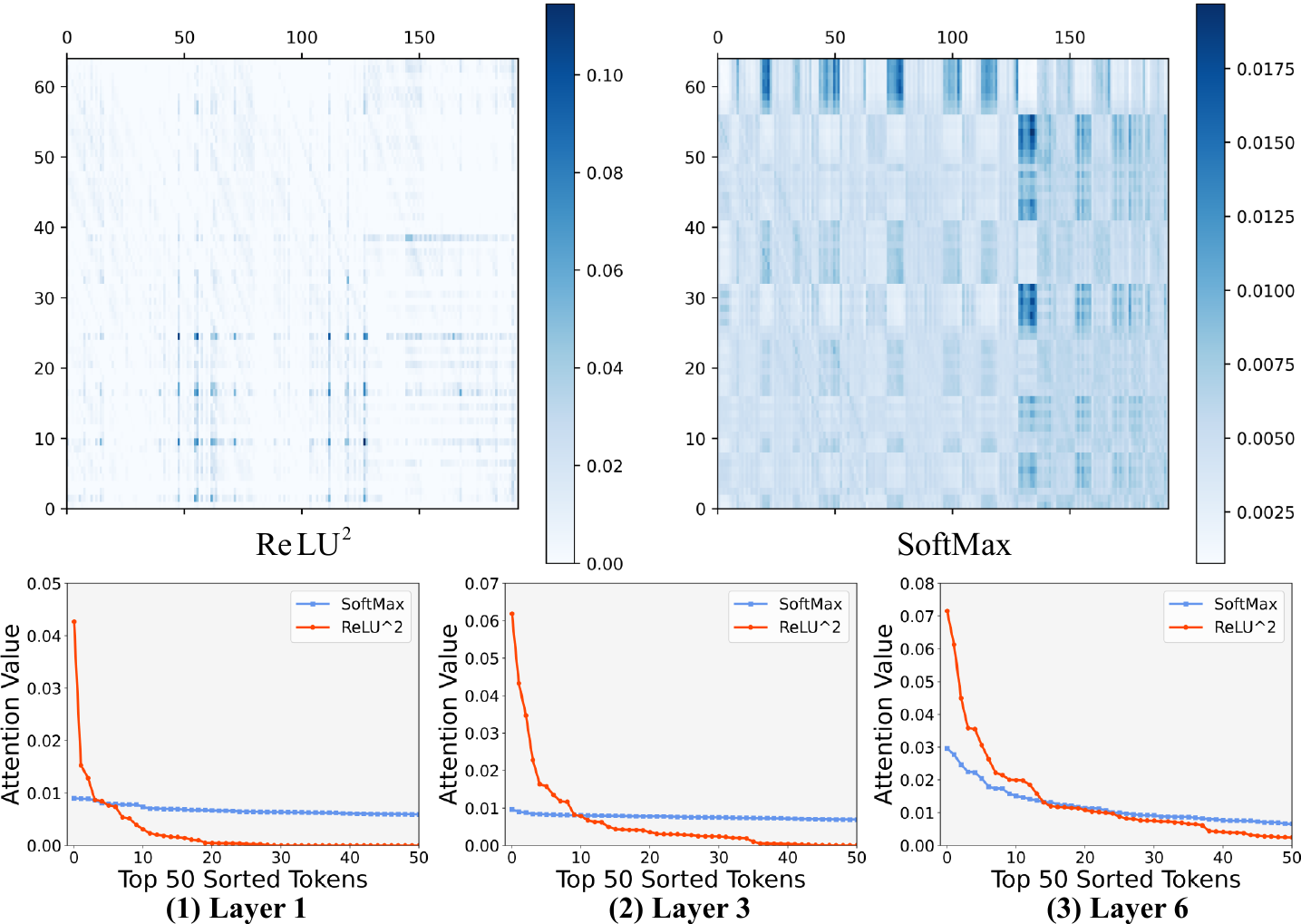}
		\caption{\textbf{Attention map comparison.} We compare the visualization of the intra\&inter attention map under $\texttt{ReLU}^2$ and the original $\texttt{SoftMax}$. The Top 50 values of all tokens (64$\times$192) of an attention map under different network depths using these two activation functions are also counted. It is obvious that $\texttt{ReLU}^2$ is significantly more focused on fewer tokens than $\texttt{SoftMax}$.}
		\vspace{-0.6cm}
		\label{fig:attentionmap}
	\end{center}
\end{figure}

\textbf{Refocused attention with sparse refocus activation.} As mentioned above, to mitigate the influence of irrelevant features in aligned hidden states, we introduce the sparse refocus activation function $\texttt{ReLU}^2$ during this refocused intra\&inter-frame attention calculation:
\begin{equation}
\label{eq:SA}
      \texttt{RITB}_\texttt{Attention} = \texttt{ReLU}^2(\bm{Q}_{m,n}^{t}{\bm{K}_{m,n}^{t}}^T/\sqrt{d}+B)\bm{V}_{m,n}^{t},
\end{equation}
where $d$ is the channel dimension of the token, and $B$ is the learnable relative positional encoding.
Unlike traditional transformer blocks that use $\texttt{SoftMax}$ to retain all feature correlations, we employ the $\texttt{ReLU}^2$ function to set negative values in the $\bm{Q} \times \bm{K}$ matrix to zero while amplifying positive correlation values.
This leads to more effective utilization of inter-frame temporal information, as illustrated in Fig. \ref{fig:attentionmap}.

\begin{figure}[!t]
	\begin{center}
		\includegraphics[width=0.36\textwidth]{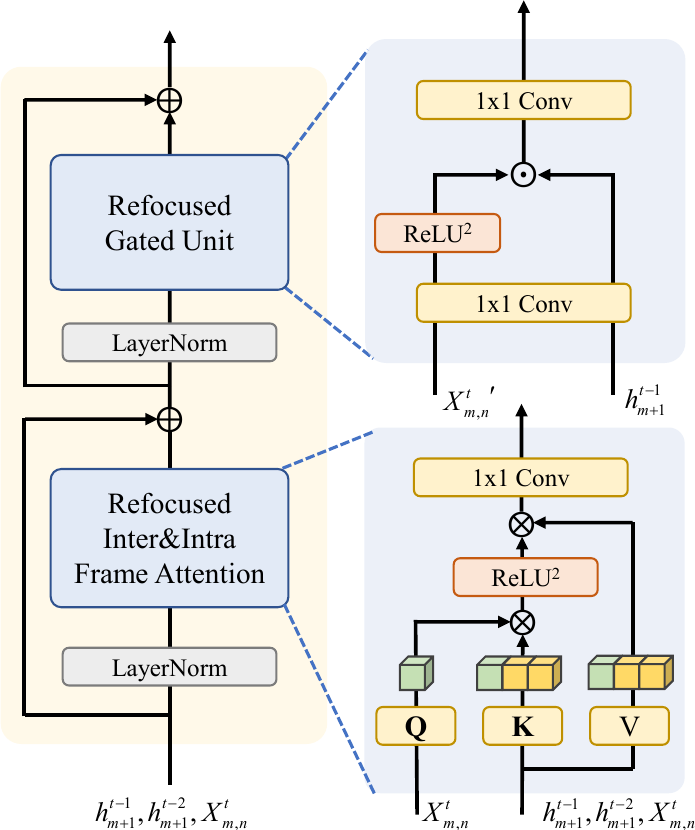}
		\caption{\textbf{Illustration of the refocused intra\&inter frame transformer block (RITB).} The refocused intra\&inter frame attention module contained in the RITB block can selectively prioritize useful temporal information. And the refocused gated unit module further facilitates the utilization of inter-frame information. More details of our RITB block can be found in Subsection \ref{sec:ritb}.}
		\vspace{-0.7cm}
		\label{fig:ritb}
	\end{center}
\end{figure}

\textbf{Refocused gated unit (RGU).}
Beyond the attention module, we further leverage inter-frame temporal information within the Feed-Forward Networks (FFN) module by introducing a \textbf{R}efocused \textbf{G}ated \textbf{U}nit (\textbf{RGU}).
Specifically, we leverage the aligned temporal propagation hidden state $\bm{h}_{m+1}^{t-1}$ of the previous frame for FFN calculation after calculating the refocused intra\&inter-frame attention to enhances the feature ${\bm{X}_{m,n}^{t}}^{\prime}$ of the current frame:
\begin{equation}
\label{eq:ffn}
      \texttt{RITB}_\texttt{FFN} = g(\texttt{ReLU}^2(f({\bm{X}_{m,n}^{t}}^{\prime}))\odot f(\bm{h}_{m+1}^{t-1})),
\end{equation}
Where $f(\cdot)$ and $g(\cdot)$are linear projections, $\odot$ is indicates element-wise multiplication and we also used $\texttt{ReLU}^2$ as the non-linear activation function.
Additionally, our transformer layer incorporates LayerNorm, a common component in Transformer-based architectures~\cite{liang2022recurrent,shi2022rethinking,xu2023implicit,Zhou_2024_CVPR}.
The overall structure of the proposed refocused intra\&inter-frame transformer block (RITB) is illustrated in Fig. \ref{fig:ritb}.

\subsection{Model Implementations}
\label{sec:MI}
In this subsection, we detail the implementation of our proposed model.
Our LRTI-VSR framework is built upon the bi-directional second-order grid propagation framework of BasicVSR++~\cite{chan2022basicvsr++}, which has also been adopted in recent state-of-the-art methods~\cite{shi2022rethinking,xu2023implicit,Zhou_2024_CVPR}.
The whole model consists of three parts, i.e. the shallow feature extraction part, the recurrent feature refinement part and the feature reconstruction part.
Generally, the recurrent feature refinement part comprises $\textbf{M}$ feature propagation modules and each feature propagation module consists of $ \textbf{N}$ cascaded processing RITB blocks.
We follow previous works~\cite{chan2022basicvsr++,shi2022rethinking,xu2023implicit,Zhou_2024_CVPR} which use a plain convolution operation to extract shallow features and adopt a pixel-shuffle layer~\cite{shi2016real} to reconstruct HR output with refined features.
The patch alignment method used in PSRT~\cite{shi2022rethinking} is used to align the hidden states with the current frame.
Following previous state-of-the-art approaches, we utilize the Charbonnier loss~\cite{charbonnier1994two} $\mathcal{L}_{sr} = \sqrt{\parallel\bm{\hat{I}}^{HQ} -\bm{I}^{HQ} \parallel^2 + \varepsilon ^2}(\epsilon=10^{-3})$ between the estimated HR image $\bm{\hat{I}}^{HQ}$ and the ground truth image $\bm{I}^{HQ}$ to train our network in all of our experiments.
The detailed overall architecture of the proposed LRTI-VSR model and structural comparison of the proposed RITB block with MFSAB and IIAB adapoted in PSRT~\cite{shi2022rethinking} and MIAVSR~\cite{Zhou_2024_CVPR} are shown in the \textbf{Supplementary Material}. Furthermore, the specific settings of our LRTI-VSR model in the various comparison experiments will be introduced in Section \ref{sec:experiments}.

\section{Experiments}\label{sec:experiments}
\subsection{Experimental Settings}
\begin{table}
% \captionsetup{font={small}}
% \scriptsize
\footnotesize
\setlength{\tabcolsep}{9pt}
\begin{center}
\begin{tabular}{l|c|cc|c}
% \toprule
\toprule
% \cmidrule(r){1-4}
\multirow{2}{*}{\textbf{Methods}} &Training& \multicolumn{2}{c}{REDS4}&GPU\\ 
&Length&PSNR& SSIM&days\\
\midrule
BasicVSR~\cite{chan2021basicvsr}     & 15& 31.42dB & 0.8909&\color[gray]{0.4}{31.2} \\
 \textit{+TB strategy}& 15&\textbf{31.56dB} &\textbf{0.9060}&35.2\\ %training strategy
 \midrule
BasicVSR++~\cite{chan2022basicvsr++} & 30& 32.39dB & 0.9069&\color[gray]{0.4}{57.6} \\
 \textit{+TB strategy}& 30&\textbf{32.49dB} &\textbf{0.9176}&60.8\\ 
 \midrule
PSRT~\cite{shi2022rethinking}  & 16& 32.72dB & 0.9106& \color[gray]{0.4}{235} \\  
 \textit{+TB strategy}& 16&\textbf{32.84dB} &\textbf{0.9234}&+21.2\\
 \bottomrule
\end{tabular}
\end{center}
\vspace{-4mm}
\caption{Ablation studies of our proposed truncated backpropagation training strategy (denoted as \textit{TB} in the table) on previous state-of-the-art CNN\&Transformer VSR models. All models are trained and evaluated on the REDS~\cite{nah2019ntire} dataset.} \label{tab: training other methods}
% \vspace{-2mm}
\end{table}

\begin{table}
% \captionsetup{font={small}}
% \scriptsize
\footnotesize
\setlength{\tabcolsep}{5.6pt}

\begin{center}
\begin{tabular}{l|cccc|cc}
% \toprule
\toprule
% \cmidrule(r){1-4}
\multirow{2}{*}{\textbf{Method}} &\multirow{2}{*}{TB}& Training &\multirow{2}{*}{Memory}&GPU& \multicolumn{2}{c}{REDS4}\\  
&&Length&&days&PSNR& SSIM\\
\midrule
\multirow{4}{*}{\textbf{Ours}}&  & 8&3.9GB &3.33& 31.08dB & 0.8987 \\
& $\surd$ & \textbf{8}&\textbf{3.8GB}& \textbf{3.75}& \textbf{31.44dB} &\textbf{0.9052}  \\  
& & 24&10.9GB& 9.12& 31.44dB & 0.9045 \\  
&  & 40&18.4GB&15.40 & \color[gray]{0.4}{31.61dB} & \color[gray]{0.4}{0.9079}  \\
  
\bottomrule
\end{tabular}
\end{center}
\vspace{-4mm}
\caption{Ablation studies on the effectiveness of our proposed truncated backpropagation training strategy (TB) in learning long-range temporal information within long video clips. All these models are trained and evaluated on one RTX4090 GPU.}  \label{tab: training strategy}
% \vspace{-2mm}
\end{table}
\begin{figure}[!t]
	\begin{center}
		\includegraphics[width=0.45\textwidth]{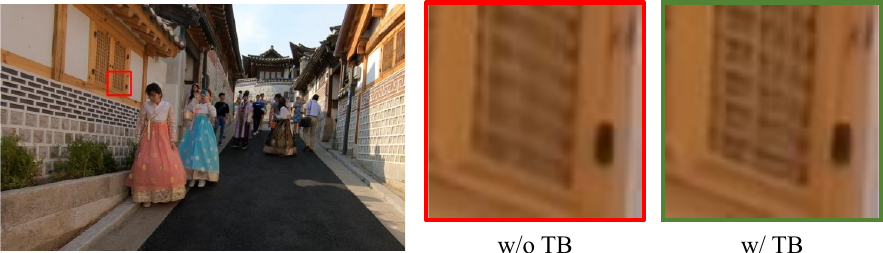}
        % \vspace{-4mm}
		\caption{\textbf{Effect of truncated back-propagation (TB) training strategy.} The contribution of our proposed training strategy is more obvious in regions with fine details. The long-term propagation information from long video clips leads to marked improvements.}
        \vspace{-4mm}
		% \vspace{-0.6cm}
		\label{fig:tbvisual1}
	\end{center}
\end{figure}
To evaluate the ability of our proposed LRTI-VSR model to learn long-range temporal dependencies within long video sequences,
we conduct extensive experiments on the REDS~\cite{nah2019ntire} and ToS3~\cite{chu2020learning} datasets.
The REDS dataset is a widely-used video dataset with each video sequence contains 100 frames, and the ToS3 dataset is the test set of 3 long video sequences (room,bridge,and face) of lengths 150, 166 and 233.
We train our LRTI-VSR model on the REDS dateset with bicubic downsampling from scratch for 600K iterations, and evaluate all the proposed VSR models on the REDS testing data (i.e. REDS4) and ToS3 datatset. %and short-video UDM10~\cite{PFNL}
In addition, we also validate our LRTI-VSR model on the real-world VideoLQ~\cite{chan2022investigating} dataset, and the experimental visual results are in the \textbf{Supplementary Material}.
We implement our model with PyTorch and train and test all of our models with RTX 4090 GPUs.
The respective hyper-parameters used for ablation study and comparison with state-of-the-art methods will be introduced in the following subsections. 

\begin{table*}
	\renewcommand{\arraystretch}{1.1}
	\renewcommand{\tabcolsep}{5.6pt}
    \begin{adjustbox}{center}
    \centering
  % \begin{tabular}{l|l}
  %\scriptsize
  %\setlength{\tabcolsep}{7pt}
    \begin{tabular}{c|c|c|c||cc|cc}%\scalebox{0.86}{
    \toprule
    % \cmidrule(r){1-4}
    \multirow{2}{*}{Method} & Training&\multirow{2}{*}{Params(M)}&\multirow{2}{*}{FLOPs(T)}& \multicolumn{2}{c}{REDS4} & \multicolumn{2}{c}{ToS3}\\
    &Length&&& PSNR & SSIM& PSNR & SSIM \\
    \midrule
    TOFlow~\cite{xue2019video} & 5& -&-& 27.98 & 0.7990&- & -\\
    EDVR~\cite{wang2019edvr} & 5&20.6& 2.95& 31.09 & 0.8800 & 32.96 & 0.9100\\
    MuCAN~\cite{li2020mucan} & 5& -&-& 30.88 & 0.8750& - & -\\%& -&- \\
    VSR-T~\cite{cao2021video} & 5& 32.6&1.60&31.19 & 0.8815& 33.11 & 0.9125\\
    VRT~\cite{liang2022vrt} & 6&35.6&1.37 &31.60 & 0.8888 & 33.68 & 0.9217\\
    \midrule
    BasicVSR~\cite{chan2020basicvsr}     & 15& 6.3& 0.33& 31.42 & 0.8909& 32.51& 0.9015\\
    IconVSR~\cite{chan2020basicvsr}      & 15& 8.7& 0.51& 31.67 & 0.8948& 33.69 & 0.9260\\
    BasicVSR++~\cite{chan2022basicvsr++} & 30& 7.3& 0.39& 32.39 & 0.9069& 34.24 & 0.9339\\%
    \midrule
    VRT~\cite{liang2022vrt}              & 16&  35.6& 1.37& 32.19 & 0.9006& 34.09 & 0.9289\\%
    RVRT~\cite{liang2022recurrent}       & 30& 10.8& 2.21& 32.75 & 0.9113& 34.47 & 0.9354\\%
    PSRT-recurrent~\cite{shi2022rethinking} & 16& 13.4& 2.39& 32.72 & 0.9106 &34.48 & 0.9361\\%
    MIA-VSR~\cite{Zhou_2024_CVPR}& 16& 16.5& 1.61& 32.78 & 0.9115 & 34.12& 0.9330\\%
    IART~\cite{xu2023implicit}& 16& 13.4& 2.51& 32.90 & 0.9138 & 34.63& 0.9386\\%
    \rowcolor{gray!20}
     LRTI-VSR(ours)& 16& 12.9& 1.54& \textcolor{red}{33.06} & \textcolor{red}{0.9162} & \textcolor{red}{34.81}& \textcolor{red}{0.9399}\\
    \bottomrule
    \end{tabular}
  \end{adjustbox}
\caption{Quantitative comparison (PSNR/SSIM) on the REDS4~\cite{nah2019ntire} and ToS3~\cite{chu2020learning} datasets for 4$\times$ video super-resolution task. The number of FLOPs(T) are computed on an LR frame size of $180\times 320$. For all experiments, we color the best performance with \textcolor{red}{red}.}  
\label{tab:sota-table}
\end{table*}
\begin{figure*}
\centering
\includegraphics[width=0.86\textwidth]{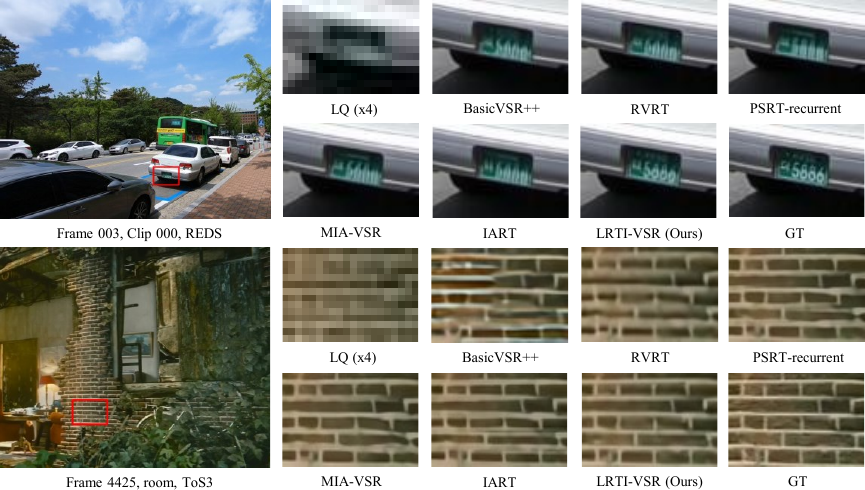}
\caption{Visual comparison for $4\times$ VSR on REDS4~\cite{nah2019ntire} dataset and ToS3~\cite{chu2020learning} dataset.}
\vspace{-0.4cm}
\label{fig:sotavisual}
\end{figure*}
\subsection{Ablation Study}
\label{sec:ablation}

\noindent\textbf{The effectiveness of proposed training strategy.}
In order to demonstrate the effectiveness of our proposed truncated backpropagation (denoted as TB) training strategy, we first apply it to several previous state-of-the-art CNN-based and Transformer-based VSR models, e.g. BasicVSR~\cite{chan2021basicvsr}, BasicVSR++~\cite{chan2022basicvsr++} and PSRT~\cite{shi2022rethinking}, respectively.
For CNN-based VSR methods BasicVSR and BasicVSR++, we follow their original basic training setups and train them using our proposed TB training strategy (truncated length 15 frames, sampling 7 times and truncated length 30 frames, sampling 5 times).
For the Transformer-based PSRT method, we perform 100K iterations on the pre-trained model using the proposed TB training strategy (truncation length 16 frames, sampling 6 times) to enhance efficiency.
The VSR results of these models on the REDS4~\cite{nah2019ntire} dataset are shown in Table \ref{tab: training other methods}.
We also report the baseline training time ({\color[gray]{0.4} gray} font) and the training time required after applying our proposed TB training strategy in this table.
With only marginally increased training costs, our proposed training strategy improves the PSNR of the CNN-based VSR models BasicVSR and BasicVSR++ by 0.14dB and 0.1dB, respectively; and for the previous state-of-the-art PSRT VSR model, our TB training strategy improves the performance of its VSR results by 0.12dB.
These results highlight the generalizability of our training strategy across different VSR architectures.

Furthermore, to evaluate the training efficiency of the proposed training strategy in learning long-term dependencies within long video sequences, we separately instantiate our LRTI-VSR model under different training strategies or length of training clips and train 300K iterations from scratch. 
We use 6 RITB blocks to to build feature propagation modules and instantiate our LRTI-VSR model with 4 feature propagation modules.
Concretely, we set training lengths of video clips as 8, 24 and 40 frames respectively to compare the efficiency of using only truncated length 8 frames with our proposed training strategy (sampling 5 times within 40 frame-length video clips).
To ensure a fair comparison, all these models are trained with batch size of 1 and evaluated on one RTX4090 GPU.
We also report the GPU memory consumption and the number of GPU days required for each experiment.
As shown in Table \ref{tab: training strategy}, it is evident that the proposed truncated back-propagation training strategy achieves a performance comparable to training with video clips of length 24 when using video clips truncated to a length of 8, while attaining an acceleration of nearly $\textbf{2.5}\times$ in training time and an reduction of nearly $\textbf{2.9}\times$ in GPU memory consumption.
Additionally, our training strategy incurs only an acceptable increase in training time compared to the baseline approach of training with video clips of length 8, but yields a substantial performance improvement of \textbf{0.36dB}.

\noindent\textbf{The effectiveness of RITB.}
In this part, we validate the effectiveness of the proposed refocused intra\&inter transformer block.
We firstly compare the proposed refocused intra\&inter transformer block (RITB) block with the multi-frame self-attention block (MFSAB) and intra\&inter-frame attention block (IIAB) which were adopted in PSRT~\cite{shi2022rethinking} and MIA-VSR~\cite{Zhou_2024_CVPR}.
As mentioned above, we also use 6 MFSAB or IIAB or RITB to build feature propagation modules and instantialize VSR models with 4 feature propagation modules and evaluate these models on one RTX4090 GPU.
\begin{table}
% \captionsetup{font={small}}
% \scriptsize
\footnotesize
\setlength{\tabcolsep}{5pt}

\begin{center}
\begin{tabular}{c|ccc|c|cc}
    \toprule
    \multirow{2}{*}{\textbf{Methods}}&\multirow{2}{*}{$\texttt{ReLU}^2$} & \multirow{2}{*}{RGU} & \multirow{2}{*}{TB}& Params & \multicolumn{2}{c}{REDS4}\\
    & & & &(M)& PSNR & SSIM \\
    \midrule
    MFSAB~\cite{shi2022rethinking}& &         &          & 6.41 & 30.76&0.8918 \\
    IIAB~\cite{Zhou_2024_CVPR}&  &          & & 6.06& 30.83& 0.8929 \\
    \midrule
    \midrule
    \multirow{4}{*}{\textbf{RITB}}&$\surd$ &  &   &6.06 & 31.02& 0.8969 \\
    & &$\surd$  &   & 6.06 & 30.97& 0.8976\\
    &$\surd$ & $\surd$ &          & 6.06 & 31.08& 0.8987\\
    &$\surd$ & $\surd$ & $\surd$ & 6.06 & \textbf{31.44}& \textbf{0.9052} \\
    
    \bottomrule
  \end{tabular}
\end{center}
\vspace{-4mm}
\caption{Ablation studies on the proposed RITB block combined with our proposed truncated backpropagation training strategy (TB). More details can be found in our ablation study subsection \ref{sec:ablation}.}
\label{tab: ritb}
\vspace{-4mm}  
\end{table}

In addition, we perform ablation experiments on the improvements proposed in RITB respectively to verify the effectiveness of each part of RITB.
As can be found in Table \ref{tab: ritb}, our model achieves a 0.19dB improvement over the baseline IIAB-VSR model with the addition of only the $\texttt{ReLU}^2$ activation function, and a 0.14dB improvement when only the Refocused Gated Unit (RGU) is added.  When combining both modifications, our LRTI-VSR model with the RITB block outperforms the IIAB-VSR model by \textbf{0.24dB} with no additional parameter increase. Furthermore, with the integration of our proposed truncated backpropagation training strategy, the LRTI-VSR model demonstrates a performance improvement by a large margin of \textbf{0.61dB} and \textbf{0.68dB} over the baseline IIAB-VSR~\cite{Zhou_2024_CVPR} and MFSAB-VSR~\cite{shi2022rethinking} model.
\subsection{Comparison with State-of-the-art Methods}
In this subsection, we compare the proposed LRTI-VSR model with current state-of-the-art VSR methods.
We compare the proposed LRTI-VSR with representative sliding-window based methods TOFlow~\cite{xue2019video}, EDVR~\cite{wang2019edvr}, MuCAN~\cite{li2020mucan}, VSR-T~\cite{cao2021video}, VRT~\cite{liang2022vrt}, RVRT~\cite{liang2022recurrent} and representative recurrent-based methods BasicVSR~\cite{chan2020basicvsr},  BasicVSR++~\cite{chan2022basicvsr++}, PSRT-recurrent~\cite{shi2022rethinking}, IART~\cite{xu2023implicit} and MIA-VSR~\cite{Zhou_2024_CVPR}; among which VRT, RVRT, PSRT-recurrent, IART and MIA-VSR are Transformer-based approaches and the other approaches are CNN-based models. 

In order to compare with state-of-the-art methods, we instantialize our LRTI-VSR model with 4 feature propagation modules and each feature propagation module contains 18 RITB blocks. 
Among them, we set the interval of skip connections to [6,6,6].
The spatial window size, head size and channel size are set to $8\times 8$, 6 and 120 accordingly.
The number of parameters in our model is on par with the recent state-of-the-art methods PSRT-recurrent~\cite{shi2022rethinking} and IART~\cite{xu2023implicit}.
In particular, we use our proposed truncated backpropagation training strategy for our LRTI-VSR model, which the truncated video clip length is set to 16 and then sampling 6 times within the whole video (100 frames, REDS).

The VSR results of different methods on the long-video test datasets can be found in Table \ref{tab:sota-table}.
Our model improves the PSNR of the baseline MIA-VSR by 0.28dB on the REDS dataset and 0.69dB on the ToS3 dataset, while having a lower number of parameters and FLOPs.
Furthermore, our model improves the PSNR of the state-of-the-art IART model by \textbf{0.16dB} on the REDS dataset and \textbf{0.18dB} on the ToS3 dataset, while the number of FLOPs (test on the REDS4 dataset, $180\times 320$ in size) is nearly \textbf{40\%} lower.
Some visual results from different VSR results can be found in Fig.\ref{fig:sotavisual}, our LRTI-VSR method is able to recover more natural and sharp textures from the input LR sequences.
%
% In our supplementary file, we also provide other instantializations of our VSR model to show the scalability of our approach and compare our method with efficient VSR models.
%
More visual examples are also provided in our in the \textbf{Supplementary Material}.

\begin{table}
    \renewcommand{\arraystretch}{1.2}
	\renewcommand{\tabcolsep}{2pt}
    \label{results-table}
    \begin{threeparttable}
    \begin{adjustbox}{center}  
    \centering
  %\begin{tabular}{l|l}
    \scalebox{0.84}{
    \begin{tabular}{c|c|c|c|c}
    \toprule
    % \cmidrule(r){1-4}
    Model&Params(M) &FLOPs(T)&Runtime(ms)  &PSNR(dB)    \\
    \midrule
    RVRT*~\cite{liang2022recurrent} &10.8&2.21 & 473& 32.75    \\
    PSRT~\cite{shi2022rethinking} &13.4&2.39 & 1041&32.72 \\
    % \rowcolor{gray!30}
    MIA-VSR~\cite{Zhou_2024_CVPR} &16.5&1.61& 822 & 32.78    \\
    IART~\cite{xu2023implicit} &13.4&2.51& 1080 & 32.90    \\
    LRTI-VSR &12.9&\textbf{1.54}& 848 & \textbf{33.06}\\
    \bottomrule
  \end{tabular}}
  \end{adjustbox}
  \begin{tablenotes} %添加此处 
	\footnotesize \item * means that uses customized CUDA kernels for better performance.
    \end{tablenotes}%添加此处
  \end{threeparttable}
  % \vspace{-4mm}
    \caption{Comparison of model size and complexity of different state-of-the-art VSR models on the REDS~\cite{nah2019ntire} dataset.}  \label{tab:running time}
    \vspace{-4mm}
\end{table}
\subsection{Complexity Analysis}
In Table \ref{tab:running time}, we report the number of parameters, the number of FLOPs, the Runtime and the PSNR on the REDS dataset by different current state-of-the art Transformer-based VSR methods.
% %
Our LRTI-VSR method has a similar or smaller number of parameters but requires the least number of FLOPs when processing video sequences.
As for the runtime, our model is not as fast as RVRT, because the authors of RVRT have implemented the key components of RVRT with customized CUDA kernels.
As the acceleration and optimization of transformers still require further research, there is room for further optimization of the runtime of our method by our relatively small FLOPs.
\section{Conclusion}
In this paper, we proposed a novel recurrent video super-resolution training framework which leverages long-range refocused temporal information, named LRTI-VSR. We develop a generic training strategy for the recurrent-based VSR model that effectively learns the accurate temporal propagation patterns within long video sequences and facilitates training using shorter video clips. Further more, a refocused intra\&inter-frame Transformer block is proposed to select and refocus features with positive contributions from the previously hidden states for current frame restoration. We evaluated our LRTI-VSR model on various benchmark video super-resolution datasets, and our model is able to achieve state-of-the-art video super-resolution results.

{
    \small
    \bibliographystyle{ieeenat_fullname}
    \bibliography{main}

\begin{thebibliography}{46}
\providecommand{\natexlab}[1]{#1}
\providecommand{\url}[1]{\texttt{#1}}
\expandafter\ifx\csname urlstyle\endcsname\relax
  \providecommand{\doi}[1]{doi: #1}\else
  \providecommand{\doi}{doi: \begingroup \urlstyle{rm}\Url}\fi

\bibitem[Beltagy et~al.(2020)Beltagy, Peters, and Cohan]{beltagy2020longformer}
Iz Beltagy, Matthew~E Peters, and Arman Cohan.
\newblock Longformer: The long-document transformer.
\newblock \emph{arXiv preprint arXiv:2004.05150}, 2020.

\bibitem[Caballero et~al.(2017)Caballero, Ledig, Aitken, Acosta, Totz, Wang, and Shi]{caballero2017real}
Jose Caballero, Christian Ledig, Andrew Aitken, Alejandro Acosta, Johannes Totz, Zehan Wang, and Wenzhe Shi.
\newblock Real-time video super-resolution with spatio-temporal networks and motion compensation.
\newblock In \emph{Proceedings of the IEEE Conference on Computer Vision and Pattern Recognition (CVPR)}, pages 4778--4787, 2017.

\bibitem[Cao et~al.(2021)Cao, Li, Zhang, Liang, and Van~Gool]{cao2021video}
Jiezhang Cao, Yawei Li, Kai Zhang, Jingyun Liang, and Luc Van~Gool.
\newblock Video super-resolution transformer.
\newblock \emph{arXiv preprint arXiv:2106.06847}, 2021.

\bibitem[Chan et~al.(2020)Chan, Wang, Yu, Dong, and Loy]{chan2020basicvsr}
Kelvin~CK Chan, Xintao Wang, Ke Yu, Chao Dong, and Chen~Change Loy.
\newblock Basicvsr: The search for essential components in video super-resolution and beyond.
\newblock \emph{arXiv preprint arXiv:2012.02181}, 2020.

\bibitem[Chan et~al.(2021)Chan, Wang, Yu, Dong, and Loy]{chan2021basicvsr}
Kelvin~CK Chan, Xintao Wang, Ke Yu, Chao Dong, and Chen~Change Loy.
\newblock Basicvsr: The search for essential components in video super-resolution and beyond.
\newblock In \emph{Proceedings of the IEEE/CVF Conference on Computer Vision and Pattern Recognition}, pages 4947--4956, 2021.

\bibitem[Chan et~al.(2022{\natexlab{a}})Chan, Zhou, Xu, and Loy]{chan2022basicvsr++}
Kelvin~CK Chan, Shangchen Zhou, Xiangyu Xu, and Chen~Change Loy.
\newblock Basicvsr++: Improving video super-resolution with enhanced propagation and alignment.
\newblock In \emph{Proceedings of the IEEE/CVF conference on computer vision and pattern recognition}, pages 5972--5981, 2022{\natexlab{a}}.

\bibitem[Chan et~al.(2022{\natexlab{b}})Chan, Zhou, Xu, and Loy]{chan2022investigating}
Kelvin~CK Chan, Shangchen Zhou, Xiangyu Xu, and Chen~Change Loy.
\newblock Investigating tradeoffs in real-world video super-resolution.
\newblock In \emph{Proceedings of the IEEE/CVF Conference on Computer Vision and Pattern Recognition}, pages 5962--5971, 2022{\natexlab{b}}.

\bibitem[Charbonnier et~al.(1994)Charbonnier, Blanc-Feraud, Aubert, and Barlaud]{charbonnier1994two}
Pierre Charbonnier, Laure Blanc-Feraud, Gilles Aubert, and Michel Barlaud.
\newblock Two deterministic half-quadratic regularization algorithms for computed imaging.
\newblock In \emph{Proceedings of 1st international conference on image processing}, pages 168--172. IEEE, 1994.

\bibitem[Child et~al.(2019)Child, Gray, Radford, and Sutskever]{child2019generating}
Rewon Child, Scott Gray, Alec Radford, and Ilya Sutskever.
\newblock Generating long sequences with sparse transformers.
\newblock \emph{arXiv preprint arXiv:1904.10509}, 2019.

\bibitem[Chu et~al.(2020)Chu, Xie, Mayer, Leal-Taix{\'e}, and Thuerey]{chu2020learning}
Mengyu Chu, You Xie, Jonas Mayer, Laura Leal-Taix{\'e}, and Nils Thuerey.
\newblock Learning temporal coherence via self-supervision for gan-based video generation.
\newblock \emph{ACM Transactions on Graphics (TOG)}, 39\penalty0 (4):\penalty0 75--1, 2020.

\bibitem[Dai et~al.(2017)Dai, Qi, Xiong, Li, Zhang, Hu, and Wei]{dai2017deformable}
Jifeng Dai, Haozhi Qi, Yuwen Xiong, Yi Li, Guodong Zhang, Han Hu, and Yichen Wei.
\newblock Deformable convolutional networks.
\newblock In \emph{Proceedings of the IEEE international conference on computer vision (ICCV)}, pages 764--773, 2017.

\bibitem[Dao et~al.(2022)Dao, Fu, Ermon, Rudra, and R{\'e}]{dao2022flashattention}
Tri Dao, Dan Fu, Stefano Ermon, Atri Rudra, and Christopher R{\'e}.
\newblock Flashattention: Fast and memory-efficient exact attention with io-awareness.
\newblock \emph{Advances in Neural Information Processing Systems}, 35:\penalty0 16344--16359, 2022.

\bibitem[Ding et~al.(2023)Ding, Ma, Dong, Zhang, Huang, Wang, Zheng, and Wei]{ding2023longnet}
Jiayu Ding, Shuming Ma, Li Dong, Xingxing Zhang, Shaohan Huang, Wenhui Wang, Nanning Zheng, and Furu Wei.
\newblock Longnet: Scaling transformers to 1,000,000,000 tokens.
\newblock \emph{arXiv preprint arXiv:2307.02486}, 2023.

\bibitem[Fuoli et~al.(2019)Fuoli, Gu, and Timofte]{fuoli2019efficient}
Dario Fuoli, Shuhang Gu, and Radu Timofte.
\newblock Efficient video super-resolution through recurrent latent space propagation.
\newblock In \emph{2019 IEEE/CVF International Conference on Computer Vision Workshop (ICCVW)}, pages 3476--3485. IEEE, 2019.

\bibitem[Gu et~al.(2021)Gu, Goel, and R{\'e}]{gu2021efficiently}
Albert Gu, Karan Goel, and Christopher R{\'e}.
\newblock Efficiently modeling long sequences with structured state spaces.
\newblock \emph{arXiv preprint arXiv:2111.00396}, 2021.

\bibitem[Isobe et~al.(2020{\natexlab{a}})Isobe, Jia, Gu, Li, Wang, and Tian]{isobe2020video}
Takashi Isobe, Xu Jia, Shuhang Gu, Songjiang Li, Shengjin Wang, and Qi Tian.
\newblock Video super-resolution with recurrent structure-detail network.
\newblock \emph{arXiv preprint arXiv:2008.00455}, 2020{\natexlab{a}}.

\bibitem[Isobe et~al.(2020{\natexlab{b}})Isobe, Li, Jia, Yuan, Slabaugh, Xu, Li, Wang, and Tian]{Isobe_2020_CVPR}
Takashi Isobe, Songjiang Li, Xu Jia, Shanxin Yuan, Gregory Slabaugh, Chunjing Xu, Ya-Li Li, Shengjin Wang, and Qi Tian.
\newblock Video super-resolution with temporal group attention.
\newblock In \emph{IEEE/CVF Conference on Computer Vision and Pattern Recognition (CVPR)}, 2020{\natexlab{b}}.

\bibitem[Isobe et~al.(2022)Isobe, Jia, Tao, Li, Li, Shi, Mu, Lu, and Tai]{isobe2022look}
Takashi Isobe, Xu Jia, Xin Tao, Changlin Li, Ruihuang Li, Yongjie Shi, Jing Mu, Huchuan Lu, and Yu-Wing Tai.
\newblock Look back and forth: Video super-resolution with explicit temporal difference modeling.
\newblock In \emph{Proceedings of the IEEE/CVF Conference on Computer Vision and Pattern Recognition}, pages 17411--17420, 2022.

\bibitem[Jo et~al.(2018)Jo, Wug~Oh, Kang, and Joo~Kim]{jo2018deep}
Younghyun Jo, Seoung Wug~Oh, Jaeyeon Kang, and Seon Joo~Kim.
\newblock Deep video super-resolution network using dynamic upsampling filters without explicit motion compensation.
\newblock In \emph{Proceedings of the IEEE conference on computer vision and pattern recognition (CVPR)}, pages 3224--3232, 2018.

\bibitem[Katharopoulos et~al.(2020)Katharopoulos, Vyas, Pappas, and Fleuret]{katharopoulos2020transformers}
Angelos Katharopoulos, Apoorv Vyas, Nikolaos Pappas, and Fran{\c{c}}ois Fleuret.
\newblock Transformers are rnns: Fast autoregressive transformers with linear attention.
\newblock In \emph{International conference on machine learning}, pages 5156--5165. PMLR, 2020.

\bibitem[Kwon et~al.(2023)Kwon, Li, Zhuang, Sheng, Zheng, Yu, Gonzalez, Zhang, and Stoica]{kwon2023efficient}
Woosuk Kwon, Zhuohan Li, Siyuan Zhuang, Ying Sheng, Lianmin Zheng, Cody~Hao Yu, Joseph Gonzalez, Hao Zhang, and Ion Stoica.
\newblock Efficient memory management for large language model serving with pagedattention.
\newblock In \emph{Proceedings of the 29th Symposium on Operating Systems Principles}, pages 611--626, 2023.

\bibitem[Li et~al.(2021)Li, Xue, Baranwal, Li, and You]{li2021sequence}
Shenggui Li, Fuzhao Xue, Chaitanya Baranwal, Yongbin Li, and Yang You.
\newblock Sequence parallelism: Long sequence training from system perspective.
\newblock \emph{arXiv preprint arXiv:2105.13120}, 2021.

\bibitem[Li et~al.(2020)Li, Tao, Guo, Qi, Lu, and Jia]{li2020mucan}
Wenbo Li, Xin Tao, Taian Guo, Lu Qi, Jiangbo Lu, and Jiaya Jia.
\newblock Mucan: Multi-correspondence aggregation network for video super-resolution.
\newblock \emph{arXiv preprint arXiv:2007.11803}, 2020.

\bibitem[Liang et~al.(2022{\natexlab{a}})Liang, Cao, Fan, Zhang, Ranjan, Li, Timofte, and Van~Gool]{liang2022vrt}
Jingyun Liang, Jiezhang Cao, Yuchen Fan, Kai Zhang, Rakesh Ranjan, Yawei Li, Radu Timofte, and Luc Van~Gool.
\newblock Vrt: A video restoration transformer.
\newblock \emph{arXiv preprint arXiv:2201.12288}, 2022{\natexlab{a}}.

\bibitem[Liang et~al.(2022{\natexlab{b}})Liang, Fan, Xiang, Ranjan, Ilg, Green, Cao, Zhang, Timofte, and Gool]{liang2022recurrent}
Jingyun Liang, Yuchen Fan, Xiaoyu Xiang, Rakesh Ranjan, Eddy Ilg, Simon Green, Jiezhang Cao, Kai Zhang, Radu Timofte, and Luc~V Gool.
\newblock Recurrent video restoration transformer with guided deformable attention.
\newblock \emph{Advances in Neural Information Processing Systems}, 35:\penalty0 378--393, 2022{\natexlab{b}}.

\bibitem[Liu et~al.(2022)Liu, Yang, Fu, and Qian]{liu2022learning}
Chengxu Liu, Huan Yang, Jianlong Fu, and Xueming Qian.
\newblock Learning trajectory-aware transformer for video super-resolution.
\newblock In \emph{Proceedings of the IEEE/CVF Conference on Computer Vision and Pattern Recognition}, pages 5687--5696, 2022.

\bibitem[Liu et~al.(2017)Liu, Wang, Fan, Liu, Wang, Chang, and Huang]{liu2017robust}
Ding Liu, Zhaowen Wang, Yuchen Fan, Xianming Liu, Zhangyang Wang, Shiyu Chang, and Thomas Huang.
\newblock Robust video super-resolution with learned temporal dynamics.
\newblock In \emph{Proceedings of the IEEE International Conference on Computer Vision (ICCV)}, pages 2507--2515, 2017.

\bibitem[Liu et~al.(2018)Liu, Saleh, Pot, Goodrich, Sepassi, Kaiser, and Shazeer]{liu2018generating}
Peter~J Liu, Mohammad Saleh, Etienne Pot, Ben Goodrich, Ryan Sepassi, Lukasz Kaiser, and Noam Shazeer.
\newblock Generating wikipedia by summarizing long sequences.
\newblock \emph{arXiv preprint arXiv:1801.10198}, 2018.

\bibitem[Nah et~al.(2019)Nah, Baik, Hong, Moon, Son, Timofte, and Mu~Lee]{nah2019ntire}
Seungjun Nah, Sungyong Baik, Seokil Hong, Gyeongsik Moon, Sanghyun Son, Radu Timofte, and Kyoung Mu~Lee.
\newblock Ntire 2019 challenge on video deblurring and super-resolution: Dataset and study.
\newblock In \emph{Proceedings of the IEEE/CVF Conference on Computer Vision and Pattern Recognition Workshops}, pages 0--0, 2019.

\bibitem[Ollivier et~al.(2015)Ollivier, Tallec, and Charpiat]{ollivier2015training}
Yann Ollivier, Corentin Tallec, and Guillaume Charpiat.
\newblock Training recurrent networks online without backtracking.
\newblock \emph{arXiv preprint arXiv:1507.07680}, 2015.

\bibitem[Pang et~al.(2021)Pang, Peng, Li, and Lu]{pang2021pgt}
Bo Pang, Gao Peng, Yizhuo Li, and Cewu Lu.
\newblock Pgt: A progressive method for training models on long videos.
\newblock In \emph{Proceedings of the IEEE/CVF Conference on Computer Vision and Pattern Recognition}, pages 11379--11389, 2021.

\bibitem[Qiu et~al.(2022)Qiu, Yang, Fu, and Fu]{qiu2022learning}
Zhongwei Qiu, Huan Yang, Jianlong Fu, and Dongmei Fu.
\newblock Learning spatiotemporal frequency-transformer for compressed video super-resolution.
\newblock In \emph{European Conference on Computer Vision}, pages 257--273. Springer, 2022.

\bibitem[Sajjadi et~al.(2018)Sajjadi, Vemulapalli, and Brown]{sajjadi2018frame}
Mehdi~SM Sajjadi, Raviteja Vemulapalli, and Matthew Brown.
\newblock Frame-recurrent video super-resolution.
\newblock In \emph{Proceedings of the IEEE Conference on Computer Vision and Pattern Recognition (CVPR)}, pages 6626--6634, 2018.

\bibitem[Shazeer(2019)]{shazeer2019fast}
Noam Shazeer.
\newblock Fast transformer decoding: One write-head is all you need.
\newblock \emph{arXiv preprint arXiv:1911.02150}, 2019.

\bibitem[Shi et~al.(2022)Shi, Gu, Xie, Wang, Yang, and Dong]{shi2022rethinking}
Shuwei Shi, Jinjin Gu, Liangbin Xie, Xintao Wang, Yujiu Yang, and Chao Dong.
\newblock Rethinking alignment in video super-resolution transformers.
\newblock \emph{arXiv preprint arXiv:2207.08494}, 2022.

\bibitem[Shi et~al.(2016)Shi, Caballero, Husz{\'a}r, Totz, Aitken, Bishop, Rueckert, and Wang]{shi2016real}
Wenzhe Shi, Jose Caballero, Ferenc Husz{\'a}r, Johannes Totz, Andrew~P Aitken, Rob Bishop, Daniel Rueckert, and Zehan Wang.
\newblock Real-time single image and video super-resolution using an efficient sub-pixel convolutional neural network.
\newblock In \emph{Proceedings of the IEEE conference on computer vision and pattern recognition}, pages 1874--1883, 2016.

\bibitem[Tallec and Ollivier(2017{\natexlab{a}})]{tallec2017unbiased}
Corentin Tallec and Yann Ollivier.
\newblock Unbiased online recurrent optimization.
\newblock \emph{arXiv preprint arXiv:1702.05043}, 2017{\natexlab{a}}.

\bibitem[Tallec and Ollivier(2017{\natexlab{b}})]{tallec2017unbiasing}
Corentin Tallec and Yann Ollivier.
\newblock Unbiasing truncated backpropagation through time.
\newblock \emph{arXiv preprint arXiv:1705.08209}, 2017{\natexlab{b}}.

\bibitem[Tao et~al.(2017)Tao, Gao, Liao, Wang, and Jia]{tao2017detail}
Xin Tao, Hongyun Gao, Renjie Liao, Jue Wang, and Jiaya Jia.
\newblock Detail-revealing deep video super-resolution.
\newblock In \emph{Proceedings of the IEEE International Conference on Computer Vision (ICCV)}, pages 4472--4480, 2017.

\bibitem[Tian et~al.(2020)Tian, Zhang, Fu, and Xu]{tian2020tdan}
Yapeng Tian, Yulun Zhang, Yun Fu, and Chenliang Xu.
\newblock Tdan: Temporally-deformable alignment network for video super-resolution.
\newblock In \emph{Proceedings of the IEEE/CVF Conference on Computer Vision and Pattern Recognition (CVPR)}, pages 3360--3369, 2020.

\bibitem[Wang et~al.(2019)Wang, Chan, Yu, Dong, and Change~Loy]{wang2019edvr}
Xintao Wang, Kelvin~CK Chan, Ke Yu, Chao Dong, and Chen Change~Loy.
\newblock Edvr: Video restoration with enhanced deformable convolutional networks.
\newblock In \emph{Proceedings of the IEEE Conference on Computer Vision and Pattern Recognition Workshops (CVPRW)}, pages 0--0, 2019.

\bibitem[Werbos(1990)]{werbos1990backpropagation}
Paul~J Werbos.
\newblock Backpropagation through time: what it does and how to do it.
\newblock \emph{Proceedings of the IEEE}, 78\penalty0 (10):\penalty0 1550--1560, 1990.

\bibitem[Williams and Zipser(2013)]{williams2013gradient}
Ronald~J Williams and David Zipser.
\newblock Gradient-based learning algorithms for recurrent networks and their computational complexity.
\newblock In \emph{Backpropagation}, pages 433--486. Psychology Press, 2013.

\bibitem[Xu et~al.(2023)Xu, Yu, Wang, Mi, and Yao]{xu2023implicit}
Kai Xu, Ziwei Yu, Xin Wang, Michael~Bi Mi, and Angela Yao.
\newblock An implicit alignment for video super-resolution.
\newblock \emph{arXiv preprint arXiv:2305.00163}, 2023.

\bibitem[Xue et~al.(2019)Xue, Chen, Wu, Wei, and Freeman]{xue2019video}
Tianfan Xue, Baian Chen, Jiajun Wu, Donglai Wei, and William~T Freeman.
\newblock Video enhancement with task-oriented flow.
\newblock \emph{International Journal of Computer Vision}, 127\penalty0 (8):\penalty0 1106--1125, 2019.

\bibitem[Zhou et~al.(2024)Zhou, Zhang, Zhao, Wang, Li, and Gu]{Zhou_2024_CVPR}
Xingyu Zhou, Leheng Zhang, Xiaorui Zhao, Keze Wang, Leida Li, and Shuhang Gu.
\newblock Video super-resolution transformer with masked inter\&intra-frame attention.
\newblock In \emph{Proceedings of the IEEE/CVF Conference on Computer Vision and Pattern Recognition (CVPR)}, pages 25399--25408, 2024.

\end{thebibliography}
}

\clearpage
\setcounter{page}{1}
\maketitlesupplementary
\appendix

In this file, we provide more implementation and experimental details which are not included in the main text. 
In Section \ref{a}, we provide a detailed diagram of the network architecture for the entire LRTI-VSR model and structural comparison of the proposed RITB block with previous work.
In Section \ref{e}, we provide additional validation experiment of the LRTI-VSR model under the real-world video dataset.
In Section \ref{b}, we provide more implementation details and more information about the dataset.
In Section \ref{c}, we provide more visual examples of our proposed LRTI-VSR model.

\section{Model Structure}\label{a}
\begin{figure*}[h]
%\captionsetup{font=small}%
%\scriptsize
%\begin{center}
% \hspace{-0.46cm}
\centering
\includegraphics[width=0.76\textwidth]{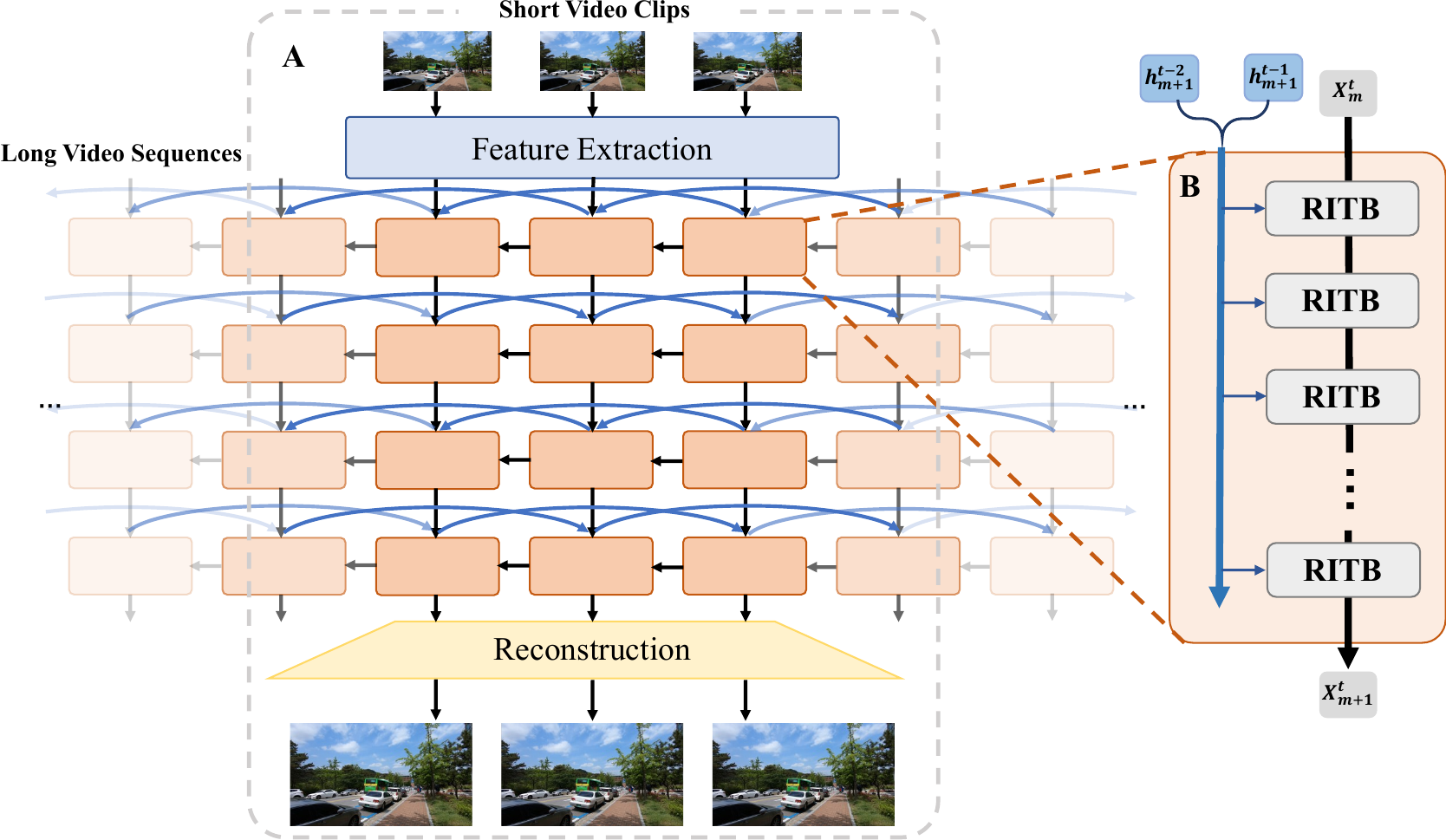}
%\end{center}%\vspace{-0.22cm}
\caption{\textbf{The overall architecture of LRTI-VSR.} 
We develop a truncated backpropagation training strategy for the VSR model (\textbf{A}) learning long-term propagation patterns within long video sequences from short video clips and further proposed a refocused intra\&inter Transformer block (RITB) used in the LRTI-VSR architecture to selectively prioritize useful information and suppress irrelevant features from temporal propagation hidden states to recover the current frame (\textbf{B}). 
}
\label{fig:arch}
\end{figure*}

\begin{figure*}[h]
\centering
\includegraphics[width=0.96\textwidth]{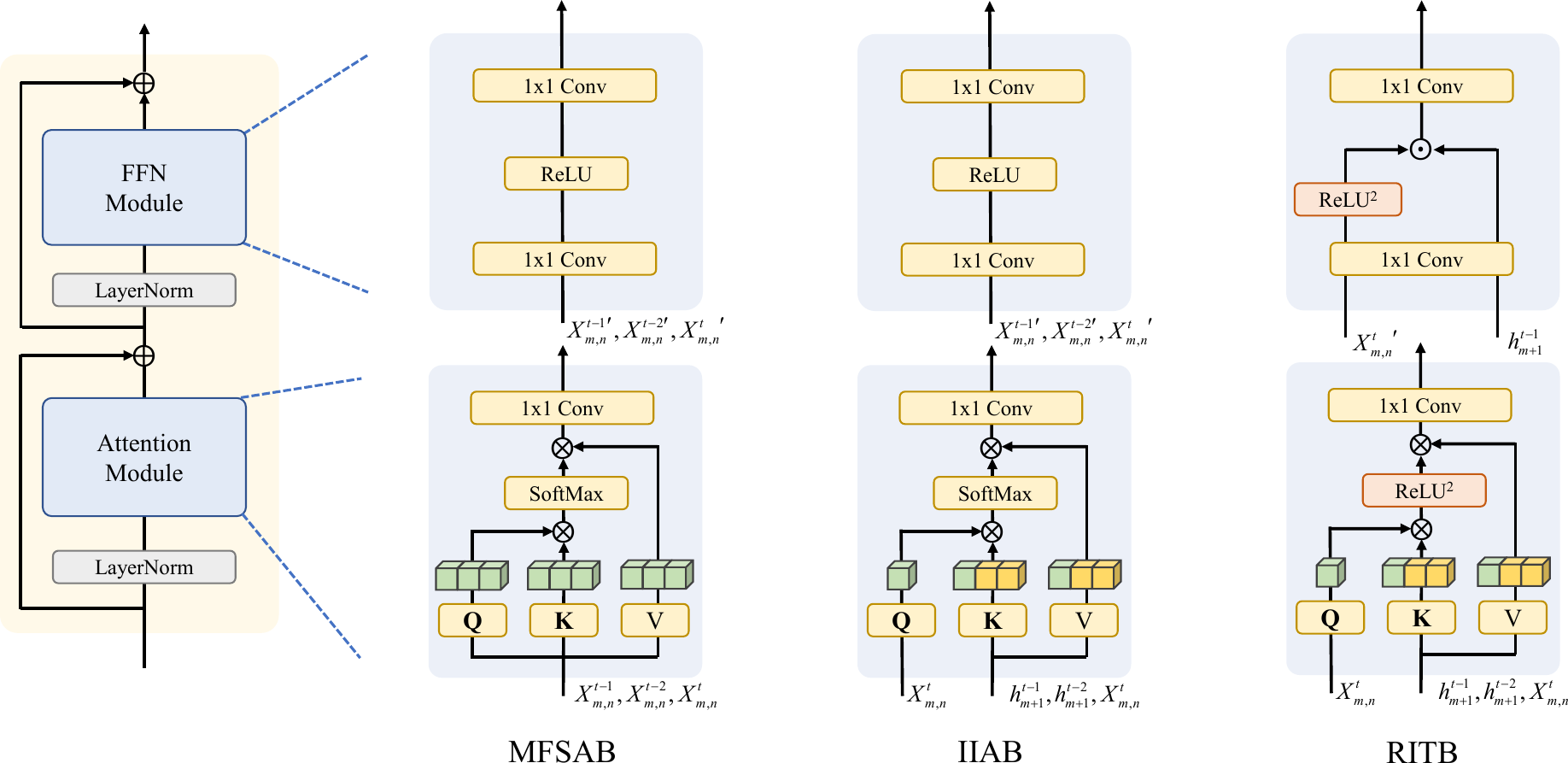}
%\end{center}%\vspace{-0.22cm}
\caption{\textbf{The structural comparison of MFSAB, IIAB and RITB.} 
Our proposed a refocused intra\&inter Transformer block (RITB) improves the previous MFSAB\cite{shi2022rethinking} block and IIAB\cite{Zhou_2024_CVPR} block in two aspects:
\textbf{(1)} We replace $\texttt{SoftMax}$ with the sparse refocus activation function $\texttt{ReLU}^2$ to suppress irrelevant features with negative impacts on the current frame’s recovery  while refocusing on beneficial features in the cross-attention module; \textbf{(2)} We integrate the aligned hidden state from previous frame into the FFN structure through a refocused gate unit, further improving inter-frame information utilization.
}
\label{fig:block}
\end{figure*}

The overall architecture of our proposed LRTI-VSR and structural comparison of the proposed RITB block with MFSAB \cite{shi2022rethinking} and IIAB \cite{Zhou_2024_CVPR} are shown in Figure \ref{fig:arch} and Figure \ref{fig:block}. We built our LRTI-VSR framework upon the bi-directional second-order grid propagation framework of BasicVSR++ \cite{chan2022basicvsr++} and the attention structure of MIA-VSR \cite{Zhou_2024_CVPR}. Besides the commonly used shallow feature extraction, recurrent feature refinement and feature reconstruction parts, we utilize the proposed truncated backpropagation training method and refocused intra\&inter Transformer block (RITB) to train and build our LRTI-VSR model.
\section{Evaluate on the Real-World Dataset}\label{e}

\begin{figure}[!t]
	\begin{center}
		\includegraphics[width=0.46\textwidth]{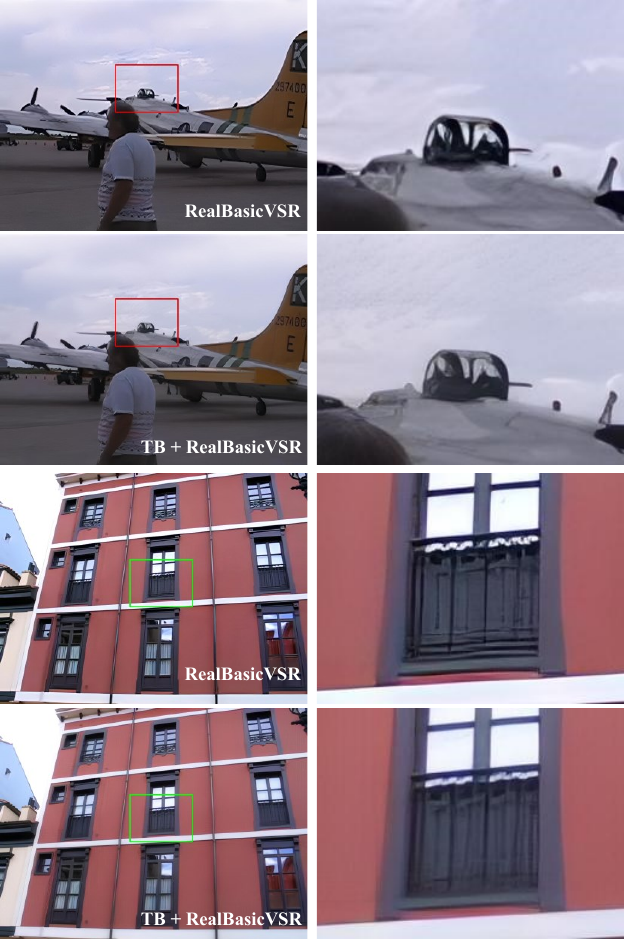}
		\caption{\textbf{Qualitative comparison on VideoLQ \cite{chan2022investigating} dataset.} Our proposed TB training strategy recovers the aircraft textures and reduces the curtain artifacts, which RealBasicVSR does not recover.}
		\vspace{-0.6cm}
		\label{fig:real}
	\end{center}
\end{figure}
% \paragraph{Evaluate on the real-world dataset.}
Real-world VSR is a variant of the VSR task where the low-resolution inputs are corrupted with different non-deterministic degradation. The classical VSR method usually has yield excessively smoothed results on this kind of dataset. To further prove that our proposed truncated back-propagation training method is also suitable for VSR models trained for real scenarios, we applied it to the RealBasicVSR \cite{chan2022investigating} model. It is obvious in the Fig. \ref{fig:real} that our training method can help produce more realistic and fine-grained results.                     

\section{Dataset and Implementation Details}\label{b}
\subsection{Datasets}
\textbf{REDS \cite{nah2019ntire}}
REDS is a widely-used video dataset for evaluating video restoration tasks.
It has 270 clips with a spatial resolution of 1280 × 720. 
We follow the experimental settings of \cite{chan2020basicvsr,chan2022basicvsr++,shi2022rethinking} and use REDS4 (4 selected representative
clips, i.e., 000, 011, 015 and 020) for testing and training our models on the remaining 266 sequences.

\vspace{3mm}
\noindent
\textbf{ToS3 \cite{chu2020learning}} ToS3 is a long video test dataset for evaluating video super-resolution tasks. It contains 3 video clips (i.e, room, bridge and face) and the length of each video clip is 150, 166 and 233 (534$\times$1280).
We follow the experimental settings of \cite{chan2020basicvsr,chan2022basicvsr++,shi2022rethinking,xu2023implicit,Zhou_2024_CVPR}  to train the LRTI-VSR model on the REDS datatset and use the 3 sequences in the ToS3 dataset to evaluate current state-of-the-art VSR models.

\vspace{3mm}
\noindent
\textbf{VideoLQ \cite{chan2022investigating}} VideoLQ is a test dataset with non-deterministic degradation for evaluating real-world VSR tasks. It contains 50 video clips and the length of each video clip is 100.
We follow the experimental settings of \cite{chan2022investigating} to train RealBasicVSR model with our proposed TB training strategy on the REDS datatset and use the VideoLQ dataset to compare with the original RealBaiscVSR model.
\subsection{{Training and Testing Details}}

\paragraph{Comparison with State-of-the-Art VSR Methods.} 
We train our LRTI-VSR model with the REDS \cite{nah2019ntire} training
dataset with zooming factor 4.
We follow the experimental settings of BasicVSR++ \cite{chan2022basicvsr++} and train our LRTI-VSR model for 600K iterations.
The initial learning rate is set as $2 \times 10^{-4}$ and a cosine learning rate decay to 1e-7. 
We train our model with Adam optimizer and the batch size is set as 24.
In the testing phase, we evaluate LRTI-VSR model's performance on the REDS4 \cite{nah2019ntire} dataset and ToS3 \cite{chu2020learning} dataset.
We calculate PSNR and SSIM on the RGB channel for these datatsets. For the calculation of the FLOPs in Table 4, we compare all VSR models with an input low-resolution (LR) frame size of 180$\times$320 on the REDS4 dataset.

\paragraph{Ablation Studies of Proposed Truncated Backpropagation Training Strategy.} 
For the ablation study of the truncated backpropagation (TB) training strategy presented in Table 1, we follow the original training settings of the BasicVSR \cite{chan2021basicvsr} and BasicVSR++ \cite{chan2022basicvsr++} models, respectively. Both models are trained from scratch using our proposed TB training strategy. For the transformer-based method PSRT \cite{shi2022rethinking}, we fine-tune the pretrained model with an initial learning rate of 1e-4 with TB training strategy. The learning rate was decayed to 1e-7 using a cosine schedule over 100K iterations with batchsize of 4.

For Table 2, we also instantiate the LRTI-VSR model with the bi-directional second-order grid propagation strategy.  The total training iterations are set to 300,000, with a learning rate initialized at 1e-4 and subjected to a cosine learning rate decay, reaching 1e-7 at the end of training. For efficiency, the batch size used for these experiments is 1, respectively. All models are trained and evaluated on the REDS dataset.

\paragraph{Ablation Studies of Proposed Refocused Intra\&inter Transformer Block (RITB).} 
For ablation studies of refocused intra\&inter Transformer block in Table 3, we instantiate the LRTI-VSR model in the same configuration as the previous ablation studies in Table 2. The total number of training iterations is also set to 300,000, with the learning rate initialized at 1e-4 and subject to cosine learning rate decay, reaching 1e-7 at the end of training.

\section{Visual Results}\label{c}
\label{morevisual}
\begin{figure*}
\centering
\includegraphics[width=0.9\textwidth]{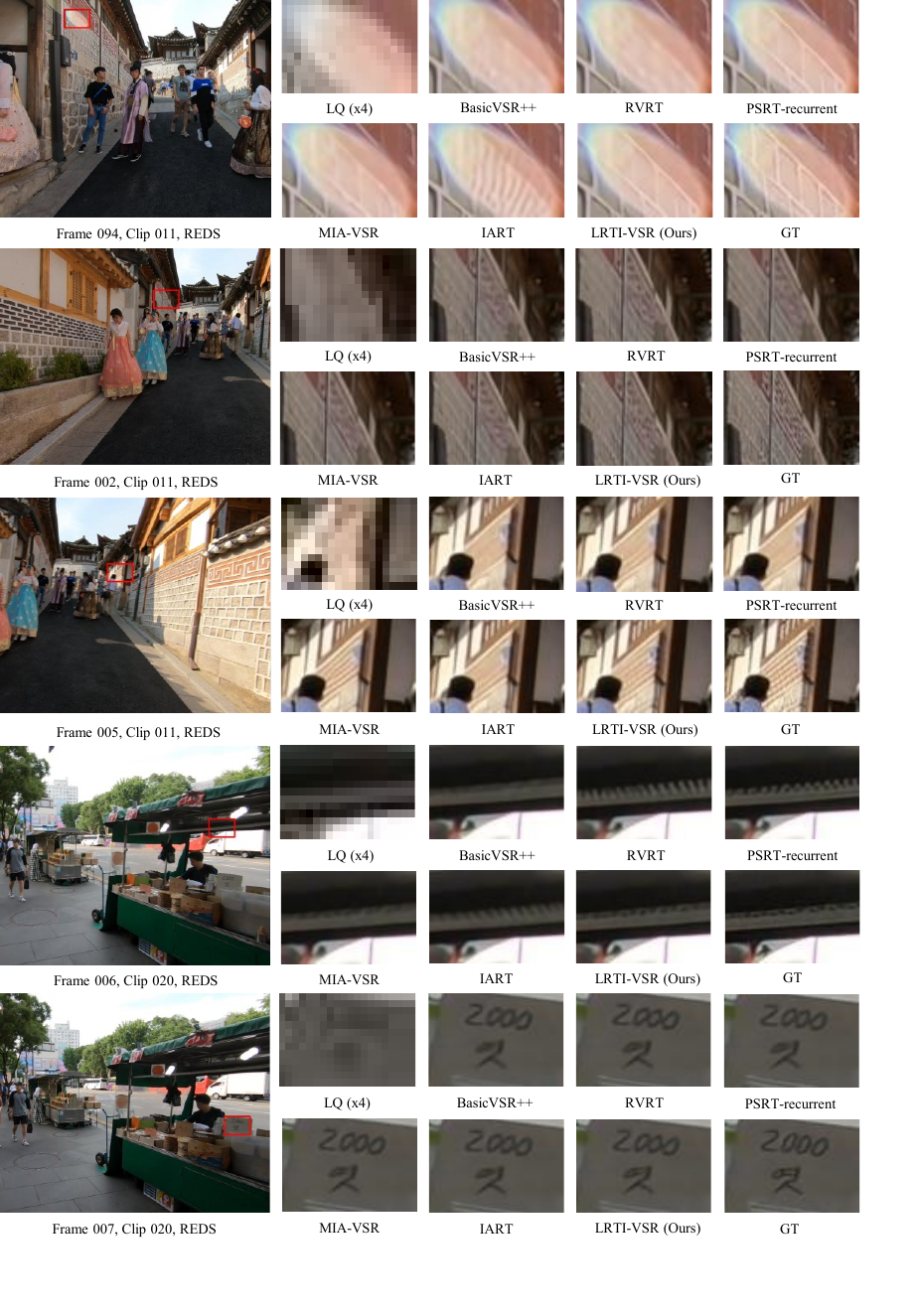}
\caption{Visual comparison for $4\times$ VSR on REDS4 dataset.}
\label{fig:morevisualreds}
\end{figure*}

\begin{figure*}
\centering
\includegraphics[width=0.9\textwidth]{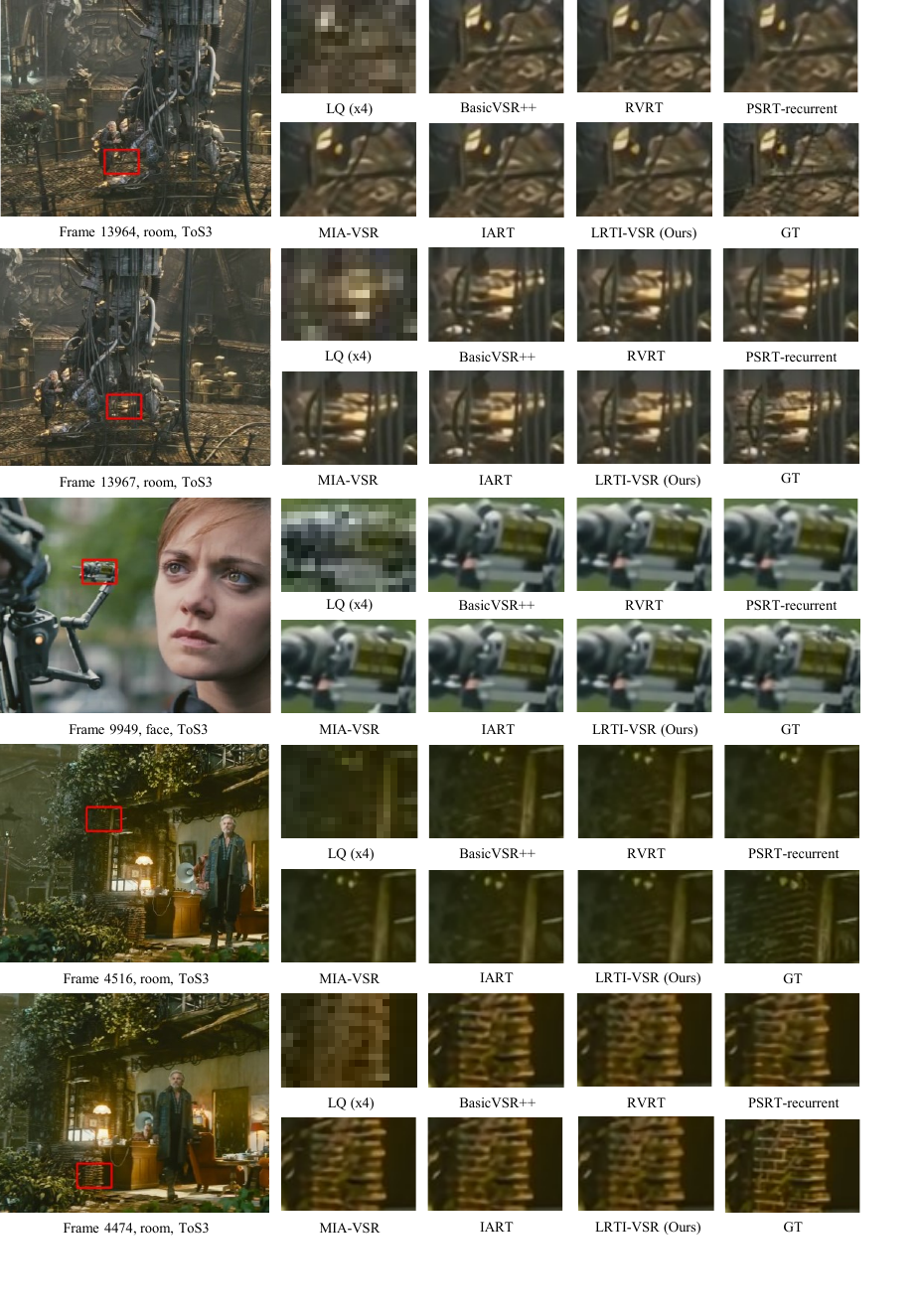}
\caption{Visual comparison for $4\times$ VSR on ToS3 dataset.}
\label{fig:morevisualvid4}
\end{figure*}

We conducted further visual comparisons between existing VSR methods and the proposed Transformaer-based recurrent video super-resolution framework utilizing long-range refocused temporal information (LRTI-VSR). The LRTI-VSR model is trained on the REDS dataset using a truncated sequence length of 16 frames. Figure \ref{fig:morevisualreds} and Figure \ref{fig:morevisualvid4} illustrate the visualization results. 
As observed, the proposed method not only improves quantitative performance but also produces images with sharp edges, fine details, and visually appealing quality, such as the edge details of buildings, license plate numbers, and intricate details in cinematic scenes.
In contrast, existing methods suffer from texture distortions or loss of detail in these scenarios.

\end{document}